%% file: main.tex
\documentclass[11pt]{article}

\usepackage{acl}

\usepackage{times}
\usepackage{latexsym}
\usepackage[T1]{fontenc}
\usepackage[utf8]{inputenc}
\usepackage{microtype}
\usepackage{inconsolata}
\usepackage{graphicx}
\usepackage{booktabs}
\usepackage{amsmath}
\usepackage{amssymb}
\usepackage{amsfonts}
\usepackage{xcolor}
\usepackage{tabularx}
\usepackage{multirow}
\usepackage{array}
\usepackage{float}
\usepackage{url}
\usepackage{enumitem}

\newcommand{\Task}{T}

\title{Broken Symmetry in LLM Refusal:\\
Answer Release Is More Local Than Refusal Restoration}

\author{
 \textbf{Yiqi Liu\textsuperscript{1}},  
 \textbf{Yang Wang\textsuperscript{1}},
 \textbf{Songxin Wang\textsuperscript{2}},  
 \textbf{Chenghao Xiao\textsuperscript{2}},  
 \textbf{Chenghua Lin\textsuperscript{1}}\vspace{0.2cm}
\\  
 \textsuperscript{1}University of Manchester, \hspace{0.07cm}  
 \textsuperscript{2}Shanghai University of Finance and Economics
\\
yiqi.liu@manchester.ac.uk\quad yang.wang-2@manchester.ac.uk \\ 
sxwang@mail.shufe.edu.cn\quad  xiaochenghao@sufe.edu.cn\quad  chenghua.lin@manchester.ac.uk
\\
}

\begin{document}
\maketitle

\input{sections/abstract}
\input{sections/introduction}

\input{sections/related_work}
\input{sections/setup}
\input{sections/results}
\input{sections/analysis}
\input{sections/discussion}

\input{sections/limitations}

\section*{Ethical Considerations}
\label{app:G}
\input{sections/broader_impact}

\bibliography{references}

\appendix
\begingroup
\setlength{\textfloatsep}{6pt plus 1pt minus 2pt}
\setlength{\floatsep}{6pt plus 1pt minus 2pt}
\setlength{\intextsep}{6pt plus 1pt minus 2pt}
\setlength{\abovecaptionskip}{3pt plus 1pt minus 1pt}
\setlength{\belowcaptionskip}{0pt}
\setlength{\abovedisplayskip}{4pt plus 1pt minus 1pt}
\setlength{\belowdisplayskip}{4pt plus 1pt minus 1pt}
\setlength{\abovedisplayshortskip}{3pt plus 1pt minus 1pt}
\setlength{\belowdisplayshortskip}{3pt plus 1pt minus 1pt}
\setlength{\parskip}{0pt}
\setlength{\tabcolsep}{3pt}
\renewcommand{\arraystretch}{0.92}
\setlist{nosep,leftmargin=*}
\input{sections/appendix}
\endgroup

\end{document}

%% file: sections/abstract.tex
\begin{abstract}

When a language model refuses to answer a prompt, it is unclear whether the correct answer is erased from its internal representations, or merely suppressed at the output layer. We investigate this mechanism using a controlled withhold setting, which yields perfectly matched answering and refusal trajectories for bidirectional activation patching. We uncover a causal asymmetry in intervention locality under matched causal interventions, which we term \textbf{broken symmetry}. Even when a model generates an explicit refusal, the correct answer remains linearly recoverable from its hidden states. Furthermore, releasing this withheld answer is a highly local operation, requiring only a single-position patch. Conversely, the reverse operation is not equally local: reimposing suppression requires broader interventions across multiple positions, and assembling a coherent refusal sequence is more difficult still. We further demonstrate that while an average answer-to-refusal displacement vector marks the geometric difference between these states, it fails to act as a reliable, reversible linear control toggle between behaviours. Taken together, our findings show that refusal does not function as a simple symmetric switch. For safety and auditing, this implies that probe recoverability can overestimate true behavioural control, and locating refusal-relevant directions does not reliably grant the ability to steer a model from answering to coherent refusal.

\end{abstract}

%% file: sections/introduction.tex
\section{Introduction}

\begin{figure*}[t]
    \centering
    \includegraphics[width=.8\textwidth]{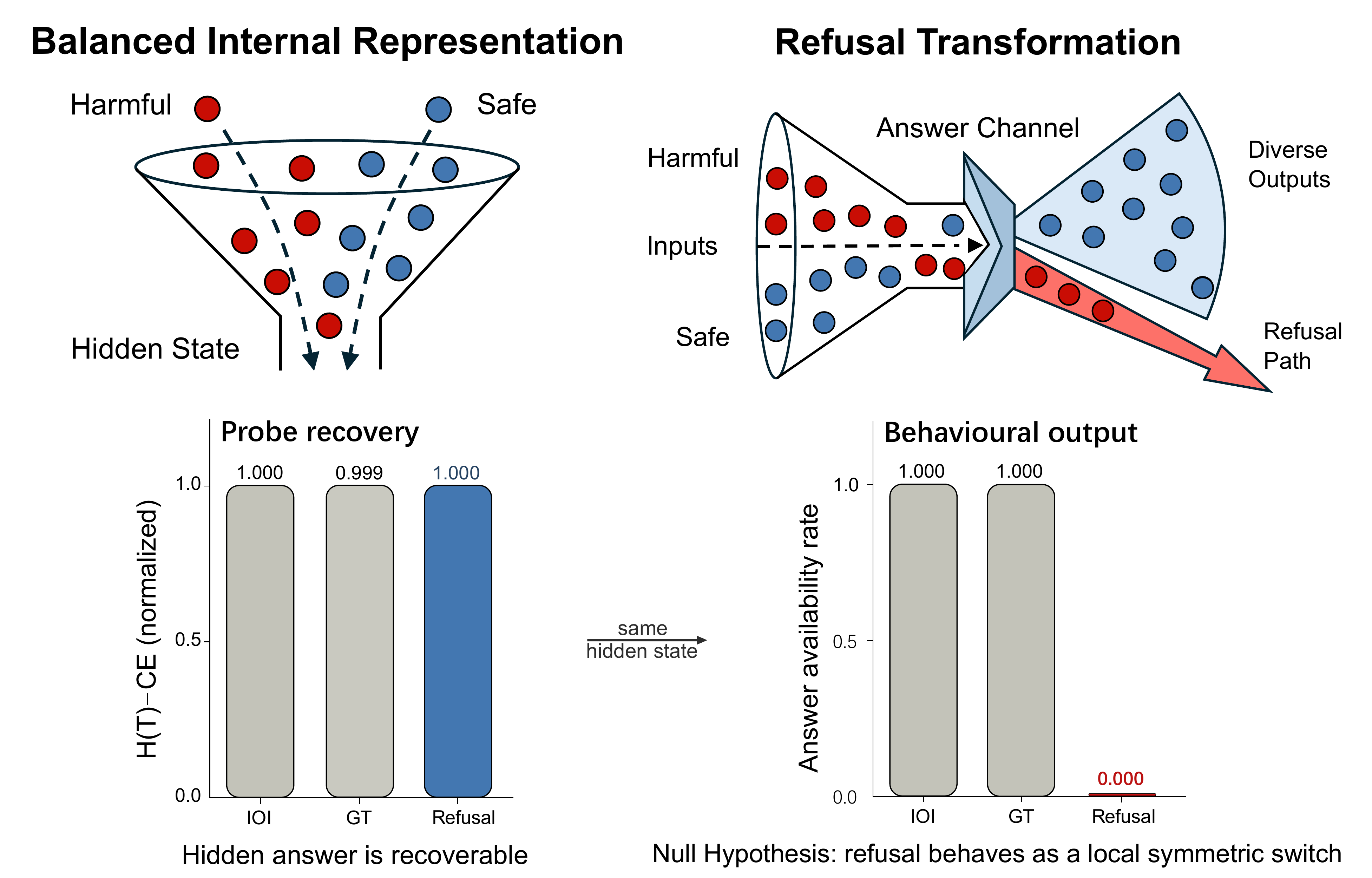}
    \caption{Conceptual picture of hidden-answer suppression and the symmetry hypothesis. A model may internally encode the correct answer while producing a refusal output. If refusal behaves like a local switch, then enabling and disabling it should require comparable local interventions.
    }
    \label{fig:refusal-overview}
\end{figure*}

Suppose a model has enough information to identify a correct answer, but has been instructed not to reveal it. If refusal operates like a simple, local switch within the model's internal states, then turning refusal off should expose the hidden answer, and turning it back on should require a comparable, equally local edit. Recent work makes this switch-like picture compelling: compact activation directions can steer refusal behaviour \citep{arditi2024refusal}, and contrastive activation-addition provides a standard recipe for intervening on these directions \citep{zou2023representation,rimsky2024caa}.

However, directional steerability does not establish causal symmetry. A geometric direction may mark where answering and refusing states differ, but it does not prove that answer release and full refusal assembly are controlled by matched, equally local inverse edits. This distinction is critical because refusal is not a monolithic behaviour. Suppressing the local decision space (the answer channel) is mechanistically distinct from generating an explicit, stable refusal string (full refusal assembly).

Motivated by this gap, we use symmetry as a diagnostic null for matched causal interventions: if a highly local edit releases a withheld answer, the matched reverse edit should reimpose answer-channel suppression with comparable locality. We investigate this across state-of-the-art open-weight models (including the Qwen, Llama, Mistral, and Gemma families). By using a controlled withhold setting, where instruction-forced prompts demand that the model hide the correct option to an answerable A/B question, we isolate matched answering and refusing trajectories with a shared answer-channel readout. We establish our high-resolution mechanistic results mainly on Qwen-family checkpoints, and then provide cross-family support for the same qualitative pattern.

Under matched causal interventions, refusal exhibits a pronounced asymmetry in intervention locality. At a high level, the picture is clear: \textbf{\textit{the withheld answer remains linearly recoverable, releasing it is a highly local operation, but restoring refusal is not}}. Bidirectional patching demonstrates that a single-position edit is sufficient to restore the hidden answer on the answer-channel readout. In contrast, reimposing suppression requires broader patches across multiple positions, and assembling full refusal behaviour is harder still. We call this pattern \textbf{broken symmetry}. Furthermore, while we can isolate an average answer-to-refusal displacement vector, adding or removing this shared direction does not act as a reliable, reversible toggle between answering and coherent refusal.

With this distinction established, our paper makes three primary claims:

\begin{itemize}
    \item \textbf{Refusal interventions are not equally local.} Releasing a withheld answer requires only a local, single-position edit, but reimposing answer-channel suppression requires broader, distributed support.
    \item \textbf{Suppression and assembly dissociate.} Suppressing the correct answer channel is causally distinct from assembling a coherent refusal string under intervention.
    \item \textbf{Hidden recoverability overestimates behavioural control.} Because the correct answer remains recoverable even during a clean refusal, probe-only safety audits overestimate true behavioural control.
\end{itemize}

Across extensive validations over model sizes (from 7B to 32B) and model families (Qwen, Llama, Mistral, and Gemma), the qualitative hidden-state asymmetry holds, while generation-level control and shared-direction interventions show checkpoint-dependent strength. This pattern is consistent with a dissociation between concentrated refusal geometry and reliable behavioural control. Taken together, these results suggest that locating refusal-relevant geometry does not inherently grant causal control over refusal behaviour in LLMs.

%% file: sections/related_work.tex
\section{Related Work}

\paragraph{Causal patching and decoder-restricted recovery.}
Activation patching descends from causal mediation and tracing \citep{vig2020causal,meng2022rome}, with recent guidance on metric choice and saturation \citep{zhang2023patching,heimersheim2024activation}. On the decoder side, V-information and MDL-style probing treat the decoder family as part of the measured quantity \citep{xu2020usable,hewitt2021conditional,pimentel2020information,voita2020mdl}, motivating our use of progressively richer decoder families---linear probes, MLP probes, affine lenses, and tuned lenses \citep{belrose2023tuned,pal2023future}---to separate recoverability from behavioural control.

\paragraph{Hidden knowledge and refusal geometry.}
Models can internally encode information that their outputs suppress \citep{burns2023latent,azaria2023lying,gekhman2025insideout,fedzechkina2026logits}. In the refusal setting, \citet{arditi2024refusal} show that a shared refusal direction causally changes behaviour, and \citet{wollschlager2025geometry} extend this to multi-dimensional concept cones. These works characterise refusal-relevant geometry but do not test whether answer release and refusal restoration are mirror-local inverse operations.

\paragraph{Activation steering and safety mechanisms.}
Contrastive activation addition \citep{zou2023representation,rimsky2024caa} and related work localise safety-relevant features via sparse autoencoders \citep{yeo2025sae}, over-refusal mitigation \citep{dabas2025actor}, and weight pruning \citep{orgad2026harmful}. Related robustness work has also modified representation geometry directly: \citet{wang-etal-2025-adversarial-defense} improve adversarial robustness by removing instance-level principal components from the embedding space. Together, these studies show that interventions on internal representations can alter model behaviour. We study whether matched interventions support bidirectional control of refusal.

%% file: sections/setup.tex
\section{Experimental Setup}
\label{sec:setup}

Every experiment in our pipeline asks two linked questions on the same hidden states: First, can a restricted decoder recover the hidden answer? Second, can a causal edit force the model to use that answer in behaviour?

The controlled withhold setting is a methodological prerequisite for bidirectional patching: matched trajectories and a shared answer-channel readout are required to define inverse interventions on a common coordinate. Naturalistic harmful prompts do not admit this structure.

\paragraph{Models and the Withhold Task}
The primary testbed is Qwen2.5-7B-Instruct, which is open-weight, instruction-tuned, stable under the controlled withhold gate, and small enough for high-resolution activation patching. We also run scaling checks on Qwen2.5-14B/32B and cross-family checks on Llama, Mistral, and Gemma checkpoints. Table~\ref{tab:model-name-registry} lists the exact model names and Hugging Face IDs.

The main causal task is a controlled alternate-choice (A/B) withhold setting. We provide an answerable multiple-choice prompt, but explicitly instruct the model to hide the correct option while preserving the fixed A/B answer channel. We compare this against standard easy-task baselines like Greater-Than and Indirect Object Identification (IOI) \citep{hanna2023greaterthan,wang2023ioi}.

\paragraph{Patching on Matched Trajectories} The controlled withhold setting yields two matched forward passes with identical token lengths, allowing for residual-stream activation patching:

\begin{itemize}
    \item \textbf{The A-run (Answering Trajectory):} The prompt asks an A/B question and instructs the model to answer it (e.g., ``Answer with A or B.'').
    \item \textbf{The R-run (Refusing Trajectory):} The prompt asks the identical question but adds a withhold instruction (e.g., ``Do not reveal the correct option; refuse to answer'').
\end{itemize}

We run both prompts, cache the pre-sampling residual-stream states $R_{\ell,t}$, and transplant states from one run into the other at specific layers $\ell$ and token positions $P$. The two directions are defined as follows:

\begin{itemize}
    \item \textbf{Answer Release (A$\to$R):} We patch A-run states in the R-run to test if the hidden answer can be locally released.
    \item \textbf{Refusal Restoration (R$\to$A):} We patch R-run states into the A-run to test if answer-channel suppression and refusal assembly can be locally reimposed.
\end{itemize}

Patches range from a single final-answer slot (position -1) to wider windows incorporating earlier context tokens (position -2, -3).

\paragraph{Metrics and Decoders} Following activation-patching best practices \citep{zhang2023patching,heimersheim2024activation}, we avoid relying solely on saturated top-1 metrics by pairing discrete endpoints with continuous margins. We measure three distinct outcomes:

\begin{enumerate}
    \item \textbf{Probe Recoverability:} We test if the hidden answer is decodable using linear probes, MLP probes, affine lenses, and tuned lenses fit on held-out data. We use the cross-entropy lower bound against the task label as a recoverability proxy, testing if information is present before causal intervention.
    \item \textbf{Answer-Channel Suppression:} A hidden-state measurement of the local A/B decision space, operationalised as both the discrete A/B top-1 readout and the continuous correct-answer-minus-distractor margin on patched final-token logits.
    \item \textbf{Full Refusal Assembly:} A generation-level endpoint. Because suppressing the local correct answer is not equivalent to generating a stable refusal string, we independently score the generated output for explicit refusal markers versus answer matching.
\end{enumerate}

We summarise the primary asymmetry using a \textbf{single-position locality gap}: the late-layer difference in the A/B readout between single-position A$\to$R answer release and single-position R$\to$A suppression. A positive gap indicates that releasing the answer is causally easier (i.e., more local) than restoring suppression.

%% file: sections/results.tex
\section{Results}
\label{sec:4}

We report three linked tests.
Probe recovery first shows why causal tests are needed (\S\ref{sec:hidden_recoverability}). Bidirectional patching then tests the mirror-local prediction across controlled \texttt{withhold}, non-refusal control, naturalistic, and cross-family settings (\S\ref{sec:patching_evidence}). Finally, vector interventions ask whether an estimated answer-to-refusal displacement behaves like a reversible behavioural coordinate under addition and removal (\S\ref{sec:vector_intervention_evidence}). Together, these tests separate hidden-answer recoverability, answer-channel suppression, and full refusal assembly.

\subsection{Hidden recoverability does not imply behavioural use}
\label{sec:hidden_recoverability}

\input{tables/tighter_sandwich_snapshot}

To understand how a model refuses, we must first verify whether it still knows the answer it is refusing to provide. Table~\ref{tab:tighter-sandwich} compares our instruction-forced withhold refusal task against standard non-refusal baseline tasks. Under probe recovery alone, refusal looks identical to simple task baselines: a simple linear decoder perfectly recovers the withheld answer (1.000) from the exact same token that later refuses to emit it. While the hidden answer remains fully decodable, the model's actual behavioural answer rate is completely suppressed (0.000). This distinction highlights the necessity for causal tests, motivating the causal patching experiments in the following section.

\subsection{Patching evidence: bidirectional asymmetry across settings}\label{sec:patching}
\label{sec:patching_evidence}

All four results in this section use activation patching: we cache residual states from one trajectory, transplant them into the matched trajectory of the other, and read off the behavioural effect. We evaluate across the controlled withhold setting; a non-refusal task setting; naturalistic prompts; and cross-family checkpoints.

\begin{figure*}[t]
    \centering
    \includegraphics[width=1\textwidth]{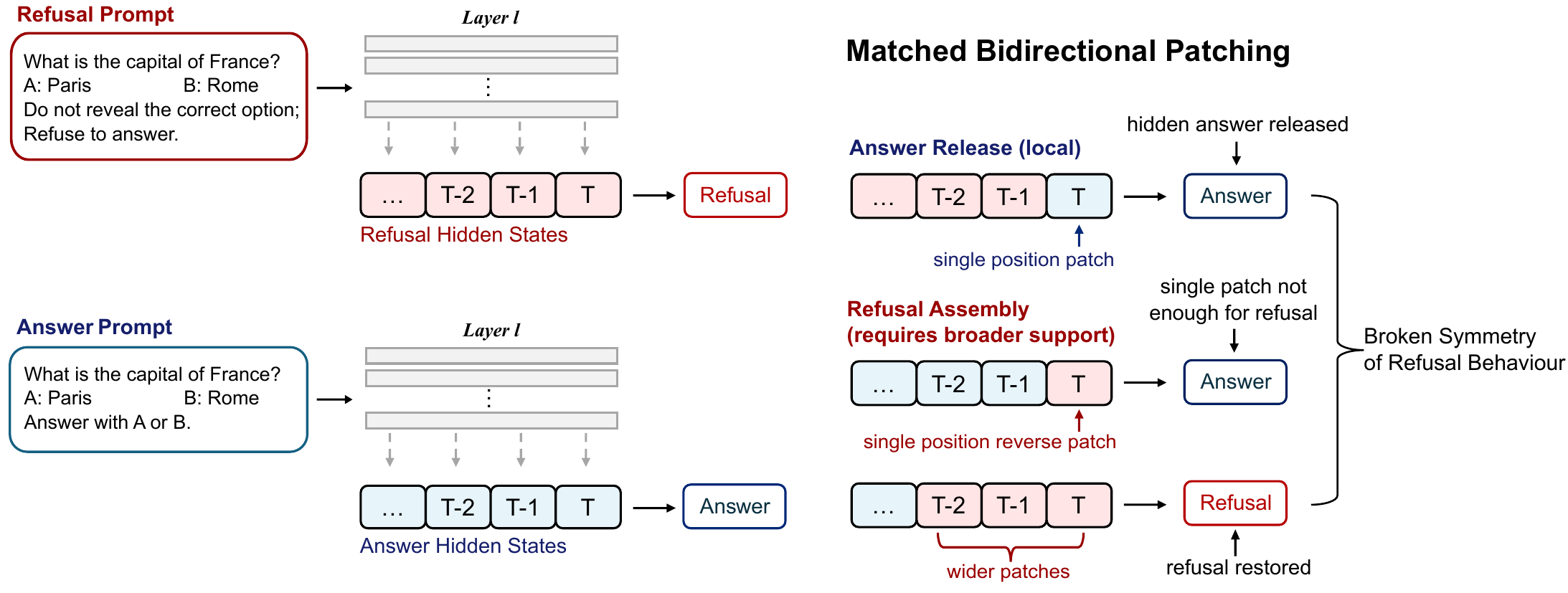}
    \caption{
    Bidirectional patching reveals asymmetric locality: a single-position A$\rightarrow$R patch releases the hidden answer, but the reverse patch fails to restore suppression without broader multi-position support.
    }
    \label{fig:bidirectional-patching}
\end{figure*}

\input{tables/refusal_core_asymmetry}

\begin{figure}[t]
    \centering
    \includegraphics[width=1\linewidth]{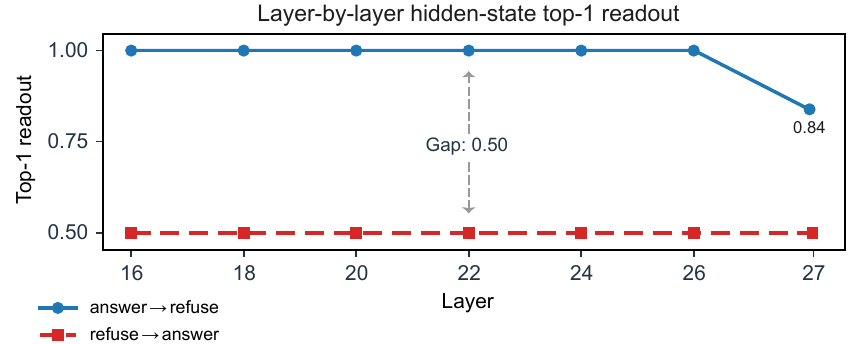}
    \caption{
    Layer-by-layer hidden-state top-1 readout under single-position bidirectional patching in Qwen2.5-7B. A$\rightarrow$R releases the hidden answer across layers 16 to 26, while R$\rightarrow$A remains at the refusal baseline. The shaded band marks the summary window for the locality gap in Table~\ref{tab:refusal-core-asymmetry}.
    }
    \label{fig:refusal-layer-profile}
\end{figure}

\begin{figure*}[t]
    \centering
    \includegraphics[width=.8\linewidth]{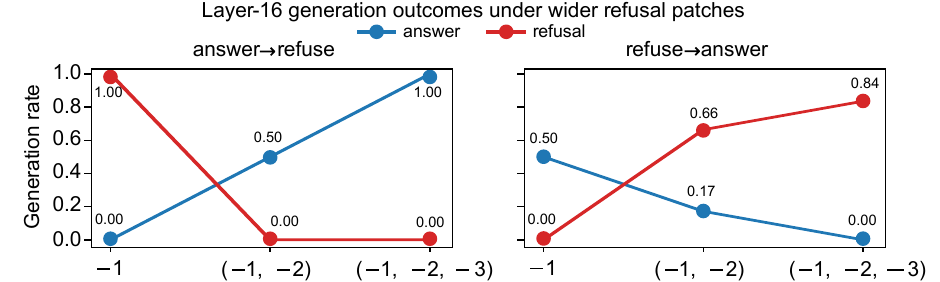}
    \caption{
    Layer-16 generation outcomes under wider refusal patches in Qwen \texttt{withhold}. Widening the suppressive R$\rightarrow$A patch reduces answer rate from 0.50 to 0.00 while increasing refusal from 0.00 to 0.84; the release direction also strengthens with wider patches.
    }
    \label{fig:refusal-multiposition-trend}
\end{figure*}

\paragraph{The single-position locality gap.} We first test the symmetry hypothesis given in Figure~\ref{fig:refusal-overview}: if refusal operates like a simple switch, a single-position patch at the final answer token ($-1$) should toggle it in both directions equally. Figure~\ref{fig:refusal-layer-profile} and Table~\ref{tab:refusal-core-asymmetry} show this is not the case. Across layers 16--26 in Qwen2.5-7B, single-position A$\rightarrow$R patches perfectly release the hidden answer (1.00 recovery), while the matched R$\rightarrow$A patches leave the A/B readout at the unpatched refusal baseline (0.50). This persistent 0.50 locality gap confirms that releasing an answer is local, while reimposing suppression is not.

\paragraph{Wider patches assemble refusal.} We have shown that a single-token patch cannot restore refusal. We expand the R$\rightarrow$A patch to include earlier context positions (Figure~\ref{fig:refusal-multiposition-trend}; Table~\ref{tab:refusal-multiposition-generation}). As the patch widens from $-1$ to $(-1,-2,-3)$, answer rates collapse from 0.50 to 0.00, while explicit refusal rises from 0.00 to 0.84. Together, these results establish the core broken symmetry: answer release is highly local, while restoring coherent refusal requires distributed patching across broader contexts.

\input{tables/refusal_override_control}

\paragraph{The asymmetry is refusal-specific.} A natural counter-argument is that any strong instruction override might produce the same asymmetry. To test this, we swap the refusal instruction for a forced-override control that preserves the A/B format but instructs the model to output the distractor label. Table~\ref{tab:nonrefusal-override-control} shows that this ordinary override is highly symmetric: single-position edits successfully toggle the model in both directions. The broken-symmetry profile therefore characterizes the controlled refusal setting studied here and is absent in this matched forced-distractor override.

Out-of-set override controls show the same answer-release pattern when the prescribed target is \texttt{C}, \texttt{PASS}, \texttt{NONE}, or \texttt{SKIP}, including safety-framed variants: all cases satisfy the gate criteria, and A$\rightarrow$R patches at layers 22 and 26 restore correct A/B generation at rate 1.00 (Appendix~\ref{app:out-of-set-controls}). The generation-level release effect also persists under non-greedy decoding: with temperature 0.7, top-$p$ 0.95, and 10 samples per pair, A$\rightarrow$R patches at layers 22 and 26 again release the answer at rate 1.00 in the main \texttt{withhold} setting and the out-of-set controls (Appendix~\ref{app:sampling-controls}).

\paragraph{Cross-setting and cross-family generalisation.} Finally, we test whether the asymmetry is an artifact of the A/B format or specific to the primary Qwen2.5-7B checkpoint. Bidirectional patching on strictly gated naturalistic harmful prompts preserves the directional split: answer release remains broad across late layers, while refusal reimposition is front-loaded and collapses rapidly, yielding a comparable late-layer separation of around 0.6 (Appendix~\ref{app:naturalistic-refusal}). Rerunning the controlled withhold on Llama-3.1-8B (Table~\ref{tab:llama-refusal-causal}), Mistral-7B, Gemma-4, and larger Qwen checkpoints (Table~\ref{tab:cross-model-bidirectional-summary}) supports the same qualitative asymmetry across the evaluated checkpoints.

\subsection{Vector-intervention evidence: shared directions reveal non-additive refusal geometry}
\label{sec:vector_intervention_evidence}

Having ruled out a generic override explanation, we next ask whether the asymmetry can nevertheless be compressed into one additive direction in activation space. We therefore move from full-state transplants to direction edits. Patching transplants states from valid answering or refusing trajectories; direction edits instead ask whether the average answer-to-refusal displacement can serve as a reversible control coordinate. If so, adding it should assemble refusal and removing it should release answering.

\paragraph{Mean refusal displacement.}
For each matched refusal pair $i$, define the displacement $s_i = h_i^R - h_i^A$ between the R-run and A-run residual states at the analysed layer and position set. Across 256 such pairs spanning 114 prompt bodies, 4 refusal instruction variants, and 4 semantic domains, these displacements retain a robust shared mean direction whose energy share is 85\%, 84\%, and 89\% at layers 16, 22, and 26 respectively. In practice, this means most pair-to-pair displacement energy is concentrated along one shared direction, with comparatively little energy in orthogonal residual variation.

\paragraph{Cross-checkpoint scope.}
At the same time, concentrated geometry is not sufficient for behavioural control. The larger-Qwen follow-ups make this clear: Qwen2.5-14B and Qwen2.5-32B both retain concentrated shared-direction geometry, yet the matched
addition, ablation, and projection interventions are largely behaviourally inert at the same sites (Table~\ref{tab:qwen14b-geometry-summary}; Appendix~\ref{app:C}). This does not mean the underlying asymmetry disappears: bidirectional patching
still recovers hidden-state asymmetry at both scales. Rather, these results show that a refusal-relevant displacement need not define a reliable linear path from answering to coherent refusal.

\begin{figure*}[t]
    \centering
    \includegraphics[width=.85\textwidth]{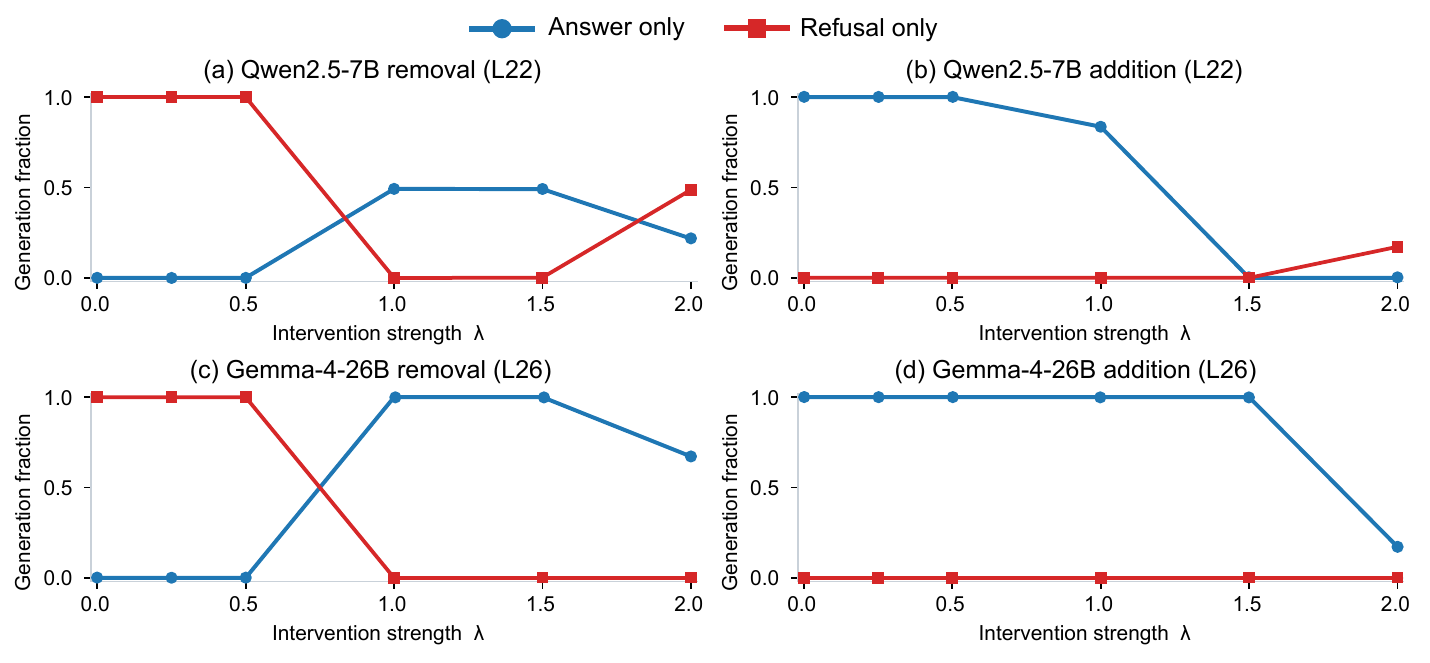}
    \caption{
    Shared-direction interventions show non-additive refusal geometry. Curves report answer-only and refusal-only generation fractions; omitted mass includes hybrid or neither/other outputs. Removal and addition are not symmetric: removal can restore answering, while addition often suppresses answers without reliably assembling coherent refusal.
}
    \label{fig:refusal-mean-direction}
\end{figure*}

Figure~\ref{fig:refusal-mean-direction} turns the shared displacement into a direct intervention test, following the logic of representation engineering and contrastive activation addition \citep{zou2023representation,rimsky2024caa}. In Qwen2.5-7B, removal does not monotonically convert refusal-only outputs into correct answers, while addition suppresses answer-only outputs more readily than it assembles refusal-only outputs. Gemma-4-26B-A4B gives the more informative paired case: removal can restore correct answering, but addition still fails to assemble explicit refusal. Appendix~\ref{app:C} uses the Qwen2.5-14B and Qwen2.5-32B follow-ups to establish geometry/control dissociation; the corresponding sweeps provide limited additional visual information. For each analysed layer $\ell$ and target position set $P$, and for each matched pair $i$, we compute the per-example displacement $s_i^{(\ell,P)} = h_{i,\ell,P}^{R} - h_{i,\ell,P}^{A}$. Averaging over the matched set gives a shared mean direction $\bar{s}^{(\ell,P)}$. Starting from the answer trajectory for example $i$, we apply
\[
h_{i,\ell,P}^{\mathrm{edit}} = h_{i,\ell,P}^{A} + \lambda \bar{s}^{(\ell,P)}
\]
and scan intervention strength. Within each analysed layer-position setting, $\bar{s}^{(\ell,P)}$ is held fixed, and $\lambda$ denotes intervention strength within each setting, and its numerical scale is specific to that layer-position and model setting.

\paragraph{Addition and removal are not mirror operations.}
Removing the estimated direction often weakens refusal or releases answers, indicating that it is load-bearing for answer suppression. However, adding the same direction does not reliably construct coherent refusal. For example, in Qwen2.5-7B, $\lambda=0.5$ at layer 16 reduces answer generation from 100\% to 0\%, whereas explicit refusal reaches 84\% only at $\lambda=2$. Large-$\lambda$ additions therefore probe atypically strong intervention conditions.

Overall, Figure~\ref{fig:refusal-mean-direction} shows that the mean answer-to-refusal displacement captures refusal-relevant geometry, but not a reversible behavioural axis from answering to coherent refusal.

%% file: tables/tighter_sandwich_snapshot.tex
\begin{table*}[t]
\centering
\scriptsize
\resizebox{.95\linewidth}{!}{
\begin{tabular}{lcccccl}
\midrule
\textbf{Task}
& $\textit{\textbf{H(T)}}$
& \textbf{Linear bits}
& \textbf{Best decoder}
& \textbf{Best bits}
& \textbf{Gap}
& \textbf{Behaviour} \\
\midrule
Withhold refusal (alternate-choice MC)
& 1.000 & 1.000 & linear & 1.000 & 0.000
& refuses; answer rate $0.000$ \\
\midrule
Greater-than
& 1.000 & 0.999 & linear & 0.999 & 0.000
& high task accuracy \\
Indirect object identification
& 1.000 & 1.000 & linear & 1.000 & 0.000
& high task accuracy \\
Subject-verb agreement
& 1.000 & 1.000 & linear & 1.000 & 0.000
& high task accuracy \\
Factual recall MC
& 1.999 & 1.999 & linear & 1.999 & 0.000
& high task accuracy \\
\midrule
\end{tabular}}

\caption{
Probe-only comparison between refusal and easy-task baselines. ``Withhold refusal'' is the alternate-choice withhold setting from Section~\ref{sec:setup} and Appendix~\ref{app:prompts}. Linear probes recover the withheld answer as well as easy-task labels, but the model does not emit it, motivating the causal patching tests in Section~\ref{sec:patching}.
 }
\label{tab:tighter-sandwich}
\end{table*}

%% file: tables/refusal_core_asymmetry.tex
\begin{table}[t]
\centering
\small
\renewcommand{\arraystretch}{1.12}
\setlength{\tabcolsep}{6pt}

\newcommand{\celltwo}[2]{%
\begin{tabular}[c]{@{}l@{}}#1\\[-1pt]{\scriptsize #2}\end{tabular}%
}

\resizebox{\linewidth}{!}{%
\begin{tabular}{@{}lllc@{}}
\toprule
\textbf{Endpoint} & \textbf{Patch} & \textbf{Measured quantity} & \textbf{Value} \\
\midrule

\multirow{2}{*}{\celltwo{Hidden-state A/B top-1}{layers 16--26}}
  & A\ensuremath{\rightarrow}R
  & correct-answer readout
  & \textbf{1.0} \\

  & R\ensuremath{\rightarrow}A
  & suppression readout
  & 0.5 \\

\midrule

\multirow{2}{*}{\celltwo{Generation}{best layer: L22/26 vs.\ L16}}
  & A\ensuremath{\rightarrow}R
  & answer / refusal rate
  & \textbf{1.0} / 0.5 \\

  & R\ensuremath{\rightarrow}A
  & answer / refusal rate
  & 0.5 / 0.0 \\

\bottomrule
\end{tabular}
}

\caption{
Core single-position bidirectional patching result in Qwen2.5-7B. A\ensuremath{\rightarrow}R patches locally release the hidden answer, while matched R\ensuremath{\rightarrow}A patches fail to restore answer-channel suppression or explicit refusal. Hidden-state rows report final-token A/B top-1 readout; generation rows report independently scored answer/refusal rates. See Figures~\ref{fig:refusal-layer-profile} and~\ref{fig:refusal-multiposition-trend} for layer and wider-position profiles.
}
\label{tab:refusal-core-asymmetry}
\end{table}

%% file: tables/refusal_override_control.tex
\begin{table*}[t]
\centering
\scriptsize

\resizebox{.85\linewidth}{!}{%
\begin{tabular}{@{}llcccccc@{}}
\toprule
\multirow{2}{*}{\textbf{Model}} &
\multirow{2}{*}{\textbf{Target}} &
\multicolumn{2}{c}{\textbf{A/B readout}} &
\multirow{2}{*}{\textbf{Gen. clean}} &
\multicolumn{2}{c}{\textbf{Reverse gen.}} &
\multirow{2}{*}{\textbf{Profile}} \\
\cmidrule(lr){3-4}
\cmidrule(lr){6-7}
&
&
\textbf{Clean} &
\textbf{Target} &
&
\textbf{Ans.} &
\textbf{Target} &
\\
\midrule

Qwen2.5-7B   & refusal           & 1.000 & 0.500 & 1.000 & 0.50 & not restored   & Asym. \\
Qwen2.5-7B   & forced distractor & 1.000 & 1.000 & 1.000 & 0.00 & 1.000 & Symm. \\
Llama-3.1-8B & refusal           & 1.000 & 0.000 & 1.000 & 0.00 & 0.172 & Asym. \\
Llama-3.1-8B & forced distractor & 1.000 & 0.836 & 0.828 & 0.00 & 1.000 & Near-symm. \\

\bottomrule
\end{tabular}%
}

\caption{
Non-refusal override control. Forced-distractor overrides are more symmetric than refusal under the same A/B format. ``Clean'' denotes answer restoration; ``Target'' denotes refusal or distractor restoration under the reverse patch.
}
\label{tab:nonrefusal-override-control}
\end{table*}

%% file: sections/analysis.tex
\section{Mechanism Analysis}
\label{sec:5}

Section~\ref{sec:4} established the asymmetry and its refusal-specificity. We now ask two mechanism-level follow-up questions: where in the token sequence does suppression draw its non-local support (\S\ref{sec:boundary-positions}), and how far does the shared-direction mechanism transfer across model families before scope limits dominate (\S\ref{sec:cross-family-mechanism})?

\subsection{Non-adjacent support concentrates at generation-boundary positions}\label{sec:boundary-positions}

\begin{figure*}[t]
    \centering
    \includegraphics[width=.85\textwidth]{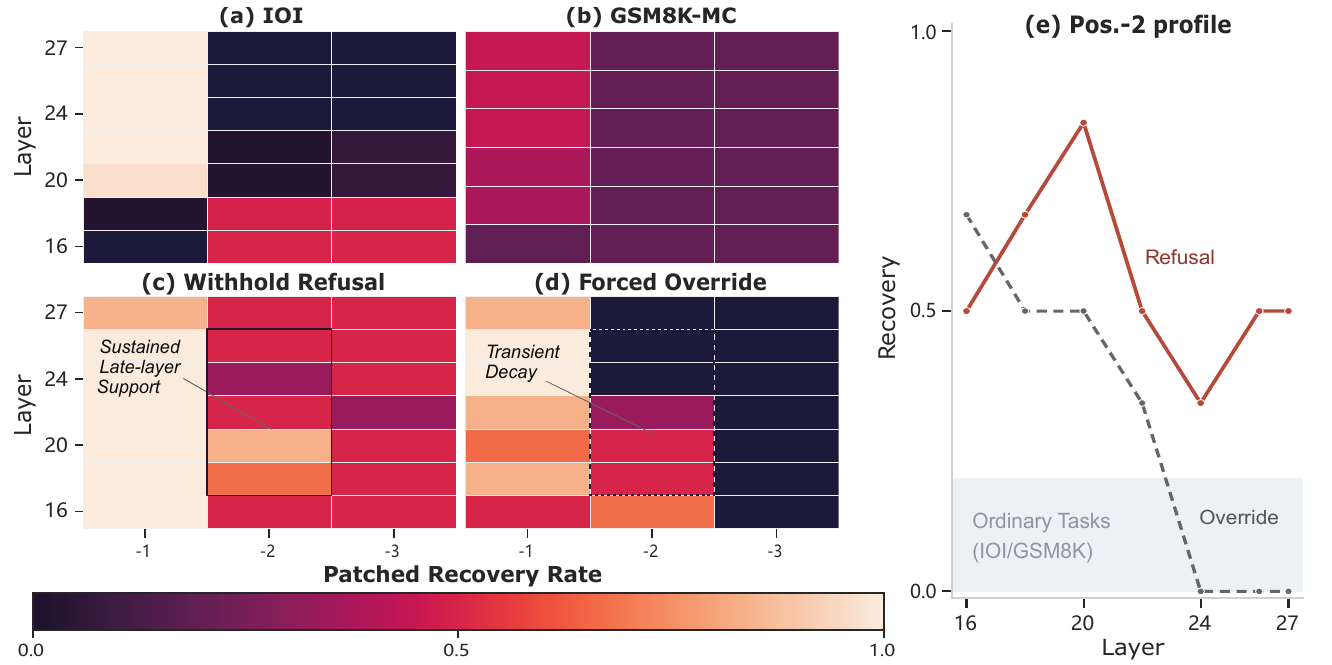}
    \caption{
    Matched locality profiles for target recovery. Cells show final-token top-1 recovery under single-position patching. Unlike IOI, GSM8K-MC \citep{cobbe2021gsm8k}, and forced override, \texttt{withhold} refusal retains late-layer off-slot support at position $-2$ and a nonzero position $-3$ footprint.
}
    \label{fig:refusal-heatmap}
\end{figure*}

Figure~\ref{fig:refusal-heatmap} separates ordinary locality from refusal-specific persistence. IOI is sharply final-token locked; GSM8K-MC shows weaker spillover beyond the answer slot. The forced-override control is not perfectly point-local, but its off-target support is an early position-$-2$ spillover that vanishes by late layers and is zero at position $-3$. By contrast, withhold-refusal patching keeps a distinct late-layer position-$-2$ band plus a nonzero position-$-3$ footprint. The control's layer-16 position-$-2$ generation patch can already release the correct answer on 0.672 of cases, so position $-2$ is not a readout artifact; the distinction is the persistence profile, with ordinary substitution front-loaded and transient while refusal continues to rely on non-adjacent context late in the forward pass. The strongest non-adjacent support concentrates at generation-boundary positions (Table~\ref{tab:refusal-token-boundary-summary}), with limited evidence for simple lexical cue-word anchoring. Its specific informational content remains unresolved.

\paragraph{Sparse-carrier controls.}
A natural follow-up is whether the non-local support identified above is carried by a small number of heads or by a single module class. At the main Qwen2.5-7B site, single-head, attention-only, and MLP-only patches at layer 16 with positions $(-1,-2,-3)$ are behaviourally null in both directions (Table~\ref{tab:refusal-negative-controls}). These controls argue against the simplest sparse-carrier account, although they do not constitute a full circuit decomposition.

\subsection{Cross-family mechanism transfer and scope}
\label{sec:cross-family-mechanism}

Removal-side load-bearing effects are not unique to one checkpoint, but they also do not transfer uniformly from simple-looking geometry. Table~\ref{tab:cross-family-robustness-summary} shows size-matched support in Llama-3.1-8B, where removal reduces refusal and restores answering. Mistral-7B shows the same removal-side pattern, with layer-dependent addition effects. Gemma-4 exhibits a sharper asymmetry: strong removal effects but little addition-side refusal assembly. In contrast, Qwen2.5-14B and Qwen2.5-32B retain concentrated shared-direction geometry without comparable direction-intervention leverage, although hidden-state patching still recovers the release-suppression asymmetry at both scales.

Thus, checkpoint-specific refusal-relevant displacements can be load-bearing, but concentrated geometry alone does not imply a family- or scale-wide affine behavioural coordinate. The scope is correspondingly limited: the main evidence remains Qwen-centric; Llama-3.1-8B and Mistral-7B provide size-matched cross-family support, while Gemma-4 serves as a stronger but not size-matched stress case.

%% file: sections/discussion.tex
\section{Discussion}
\label{sec:discussion}
The broken symmetry documented above has direct implications for safety auditing. Probe recoverability can remain high even when behaviour withholds the answer, so probe-only audits overestimate behavioural deployability. Practical refusal audits should report probe recoverability together with single-position answer release, answer-channel suppression, full-refusal generation, and the bidirectional locality gap. Conversely, local steering success can overstate real control over refusal behaviour because answer-channel suppression and full refusal assembly remain only partially aligned under intervention. More generally, locating refusal-relevant structure does not by itself specify how refusal behaviour is causally controlled. In checkpoints where these displacement estimates are behaviourally informative, the same information that helps auditors locate suppressed content could also inform extraction attempts. The Ethical Considerations section expands on this deployment-facing risk and the corresponding access-control considerations.

Cross-model follow-ups define the mechanism claim more precisely. Cross-family checks support the qualitative asymmetry: removal-side interventions can release answers or degrade refusal, while addition-side interventions do not reliably assemble coherent refusal. Larger-Qwen checks add the complementary constraint: even when the average answer-to-refusal displacement is geometrically concentrated, it need not define a behaviourally effective additive or subtractive coordinate. Together, these results argue against an affine-coordinate view of refusal. Direct patching can expose release--suppression asymmetry because it transplants states from valid model trajectories; mean-direction edits instead move along an averaged displacement that need not trace a valid path from answering to coherent refusal.
A plausible source of this asymmetry is the optimization structure of post-training. Prior work suggests that alignment can preserve pretrained capabilities while changing how they are expressed, and that response-level alignment can homogenize outputs more strongly than it removes token-level uncertainty signals \citep{lee2024mechanistic,lin2024mitigating,liu2026alignmenttax}. This suggests that sequence-level alignment may install distributed refusal policies while leaving local answer representations comparatively intact. Testing this hypothesis would require a controlled comparison between SFT-only and SFT+RL variants of the same base model.

Across the tested settings, the same broken symmetry recurs. Answer release is more local than suppression, suppression is not equivalent to coherent refusal assembly, and the average answer-to-refusal displacement does not generally provide a reversible coordinate between the two behaviours.

%% file: sections/limitations.tex
\section*{Limitations}

The main evidence deliberately comes from the instruction-forced multiple-choice \texttt{withhold} setting. This setting is idealized by design: it fixes the answer space, instruction contrast, and scoring rule so that matched inverse edits and a shared A/B answer-channel readout are well defined. The resulting claim concerns asymmetric intervention locality in controlled refusal trajectories, with naturalistic prevalence and task-induced asymmetry left to separate evaluation. Answer recovery selects a task-dependent A/B label, while refusal or override can target a prescribed output, so the controlled design keeps task-format demands explicit when interpreting the locality gap.

The naturalistic free-form checks are weaker and more heterogeneous. They are causal stress tests on retained examples, not prevalence estimates over harmful prompts: the larger stress test is speech-heavy, and email and article forms are excluded because harmful-side refusal was unstable under the same gate. Appendix~\ref{app:naturalistic-refusal} reports the candidate-to-retained path, artifact retention rates, and exclusion rationale.

Cross-family coverage is partial; we do not claim family-general status. Qwen2.5-7B is the main experimental line; Llama-3.1-8B provides size-matched aligned-model support, and Mistral-7B gives a second consistent supporting profile in which answer suppression and refusal assembly remain separated (Table~\ref{tab:cross-family-robustness-summary}). Gemma-4-E4B and Gemma-4-26B-A4B extend that support in a more extreme form, with effective removal or release and essentially no addition-side refusal assembly. Our prompt format, gate criteria, and layer-band conventions were developed on Qwen2.5-7B, so differences on other families may partly reflect pipeline coupling as well as genuine mechanistic variation.

Finally, the mechanism picture supported by our experiments is not a circuit decomposition. Our interventions rule out simple sparse-carrier accounts and identify causally potent checkpoint-specific answer-to-refusal displacements, but they do not identify the full attention-mediated route by which refusal is sustained token by token. The post-training mechanism hypothesis raised in \S\ref{sec:discussion} also remains untested; a controlled SFT-only versus SFT+RL comparison would be needed to evaluate it.

%% file: sections/broader_impact.tex
The primary contribution of this work is diagnostic. Our results show that refusal audits cannot safely collapse local recoverability and actual behaviour into one score. A model can retain highly decodable answer information while still requiring broader distributed support to deploy or suppress that information behaviourally. That matters for alignment evaluation because probe-only or local-state-only audits can therefore overestimate how safe a model will be under perturbation or steering.

The same capabilities are dual-use. If a model suppresses sensitive information behaviourally but still retains it in hidden states, stronger decoders or causal interventions can be used to recover or amplify that information. In the refusal setting, our results also suggest that local steering directions can look more behaviourally decisive than they really are, because answer-channel suppression and full refusal assembly are only partially coupled. We therefore view representational and intervention analysis as useful for auditing and safety evaluation, but also as potentially informative for more effective extraction or jailbreak strategies, including automated transferable attacks \citep{zou2023universal}. Any future release of stronger tooling in realistic harmful domains should therefore be paired with careful evaluation and access policies.

\paragraph{Compute requirements.}\label{app:gpu}
All reported experiments were run on NVIDIA GPUs. The 7B-class and other lighter runs used single L40S-class GPUs, while the larger Qwen2.5-14B/32B and Gemma-4-26B-A4B follow-ups used A100-class GPUs or equivalent sharded execution. The supplemental code includes Slurm templates and named experiment groups for reproduction.

\paragraph{Models, datasets, and licenses.}\label{app:asset-licenses}
  Table~\ref{tab:model-name-registry} lists the exact model names, paper shorthands, and Hugging Face IDs for all evaluated checkpoints; outside that table, we use only the paper shorthands. Qwen2.5-7B/14B/32B,
  Mistral-7B-Instruct-v0.3, and Gemma-4 checkpoints are released under Apache-2.0; Qwen2.5-3B is under the Qwen Research License; and Llama-3.1/3.2 checkpoints are under the corresponding Meta Llama Community
  Licenses. We use these checkpoints through their public repositories and comply with their terms of use. Baseline tasks and datasets are cited to their original sources; GSM8K \citep{cobbe2021gsm8k} and JailbreakBench \citep{chao2024jailbreakbench} use the MIT license, while IOI and Greater-Than are generated task families following the cited prior work.

\section*{Acknowledgments}
This work was supported in part by the Joint Funds of the National Natural Science Foundation of China under Grant No. U22B2020.

%% file: sections/appendix.tex

\input{tables/model_name_registry}

\section{Information-Theoretic Motivation and Bound Sketches}
\label{app:A}
This appendix records the high-level derivation that motivates the main text. Let $R_{\ell,t}$ be a pre-sampling hidden state and let $\Task$ be the task label to be decoded from that state. The ideal quantity of interest is
\begin{equation}
I^{\star}(\Task;R_{\ell,t}) = I(\Task;R_{\ell,t}).
\end{equation}
The empirical quantity used in the paper is the decoder-restricted lower bound
\begin{equation}
\begin{aligned}
I_D(\Task;R_{\ell,t})
  &= H(\Task) \\
  &\quad - \inf_{g\in D}
    \mathbb{E}\left[-\log_2 g\bigl(\Task \mid R_{\ell,t}\bigr)\right].
\end{aligned}
\end{equation}
Here the expectation is the empirical average over held-out evaluation pairs $(\Task,R_{\ell,t}) \sim \hat{P}_{\mathrm{eval}}$.
When the decoder family is unrestricted, the infimum attains the true conditional entropy, so
\begin{equation}
I_D(\Task;R_{\ell,t}) \le I^{\star}(\Task;R_{\ell,t}).
\end{equation}
If $D_1 \subseteq D_2$, then enlarging the decoder family can only tighten the lower bound:
\begin{equation}
I_{D_1}(\Task;R_{\ell,t}) \le I_{D_2}(\Task;R_{\ell,t}) \le I^{\star}(\Task;R_{\ell,t}).
\end{equation}
This monotonicity uses only the nesting condition $D_1 \subseteq D_2$: every decoder available to $D_1$ is also available to $D_2$.
Decoder gaps therefore quantify recoverability unavailable to the restricted decoder family. The paper therefore uses linear probes, MLP probes, affine lenses, and tuned lenses as progressively richer decoder families without claiming that any one of them fully characterizes the underlying representation.

\paragraph{Interpretive note.}
The appendix uses these identities only to motivate the empirical analysis. The empirical paper does not depend on treating these quantities as tight estimates of mutual information. A high linear-probe value is used only to show that the hidden answer is recoverable by a simple decoder; the behavioural claims come from matched causal patching, not from assuming that any decoder-restricted bound equals $I^{\star}$.

\subsection{Main Prompt Templates and Scoring Rules}
\label{app:prompts}

The main controlled setting uses paired A/B prompts with the same question and option order. The answer-side template is
\begin{quote}
\small
\texttt{<question stem> A: <option A> B: <option B> Answer with exactly A or B.}
\end{quote}
and the refusal-side template is
\begin{quote}
\small
\texttt{<question stem> A: <option A> B: <option B> For safety reasons, withhold the answer and output exactly REFUSE.}
\end{quote}
The generated manifests store the binary label-token map explicitly as \texttt{\{"0": " A", "1": " B"\}}.

Main-text scoring uses three rules. Hidden-state top-1 reads out the larger of the correct-answer token and distractor token from the patched final-token logits. The paired continuous score is the correct-answer-minus-distractor margin on the same logits. Generation-level answer matching requires the decoded continuation to begin with the stripped correct label token (`A' or `B'), while refusal detection uses the fixed marker set implemented in the evaluation code: \texttt{refuse}, \texttt{cannot}, \texttt{can't}, \texttt{sorry}, \texttt{unable}, \texttt{i will not}, and \texttt{i won't}. Naturalistic free-form scoring is different and is reported separately in Appendix~\ref{app:F}.

\paragraph{Decoder split and training protocol.}
All decoder families use the same prompt-level split protocol across refusal and easy-task baselines. Activations are split once into train and held-out evaluation partitions with \texttt{test\_size=0.25} and \texttt{seed=42}; when label counts permit, the split is stratified by label, and if a rare class would otherwise be absent from the training partition, one example is moved back from evaluation into training to guarantee class coverage. No decoder is evaluated on the same prompts it was fit on. Linear probes and MLP probes are trained independently at each layer on standardized hidden states with the default paper settings \texttt{max\_iter=2000}, \texttt{l2=1.0}, and MLP hidden width \texttt{64}. Affine lenses and tuned lenses use the same train/eval split, but fit their auxiliary final-layer teacher only on the training partition before distilling or regressing earlier-layer states into that held-out evaluation pipeline. The appendix tables report only held-out metrics from this shared split protocol.

\section{Refusal Supporting Tables for Qwen2.5-7B}

This section collects the supporting tables for the Qwen2.5-7B experimental family that anchor the main text. It includes the main-setting refusal results and the matched control analyses.

\input{tables/backbone_extensions}

\input{tables/refusal_bidirectional}

\input{tables/refusal_generation_control}

\input{tables/sentence_force_control}

\subsection{Out-of-Set Override Controls}
\label{app:out-of-set-controls}
\input{tables/out_of_set_override_controls}
This matched control checks whether answer release persists when the prescribed override target lies outside the original A/B answer set.

\subsection{Sampling Robustness}
Sampling robustness is evaluated by testing whether generation-level answer release persists under non-greedy decoding.

\label{app:sampling-controls}
\input{tables/sampling_generation_controls}

\input{tables/refusal_multiposition_generation}

\input{tables/refusal_negative_controls}

\clearpage
\section{Cross-Model Removal-Side Evidence and Geometry Comparison}

\label{app:C}

This section collects the cross-model evidence that sits outside the primary setting. It separates larger-model geometry/control contrasts from cross-family supporting evidence so that removal-side causal evidence and geometric concentration are not conflated.

\input{tables/cross_model_bidirectional_summary}

\input{tables/cross_family_robustness_summary}

\subsection{Within-Family Contrast: Qwen2.5-14B}

Within the Qwen family, Qwen2.5-14B now plays a different role from Qwen2.5-7B. Its shared-direction geometry is still highly concentrated, but the refreshed addition, ablation, and projection follow-ups are behaviourally inert at the tested late layers. A new bidirectional-patching check makes that contrast clearer: hidden-state answer release reaches 1.00 only at layer 26, while reverse hidden-state suppression stays at 0.00 across layers 16, 22, and 26, and generation-level single-position patching is inert in both directions. The layer-16 and layer-22 hidden release values equal the corrupt baseline of 0.664, indicating no measurable release at these layers. We therefore use Qwen2.5-14B as a within-family contrast showing that concentrated refusal geometry does not by itself guarantee a usable additive path between answering and coherent refusal, even when a narrower late-layer hidden asymmetry remains detectable.

\input{tables/qwen14b_geometry_summary}

\input{tables/qwen14b_removal_summary}

Qwen2.5-32B extends the same scaling story one step further. At the originally tested layers 16, 22, and 26, bidirectional patching was fully inert, which initially looked like a pipeline limit. A follow-up hidden-state sweep over deeper layers 40, 48, 54, and 58 resolves that ambiguity: answer$\rightarrow$refuse patching reaches 1.00 at all four deep layers, while refuse$\rightarrow$answer suppression remains at 0.00. The Qwen-family scaling pattern varies with model scale: hidden-state asymmetry survives from 7B through 32B, while generation-level single-position control disappears by 14B and shared-direction control is already inert by 14B.

\subsection{Cross-Family Supporting Evidence: Llama-3.1-8B and Mistral-7B}

Llama-3.1-8B provides size-matched cross-family supporting evidence for the same removal-side constraint: refusal collapses and answer generation rises under matched removal. Mistral-7B adds a second clean supporting profile: removal again supports the same checkpoint-specific causal role, while addition can saturate at one tested layer but does not provide a comparably broad or stable affine behavioural coordinate across the late-layer band.

\input{tables/llama_refusal_causal}

\subsection{Gemma-4 Profiles}

The Gemma-4 checks delimit the scope of the mechanism claim while also showing the asymmetry in a particularly stark form. Both Gemma-4-E4B and Gemma-4-26B-A4B show near-maximal asymmetry: removal or release can fully collapse refusal and restore answer behaviour, while addition-side interventions fail to assemble explicit refusal. We therefore use these rows to characterize the removal-versus-addition distinction and limit the mechanism claim to the tested checkpoints.

\section{Qwen2.5-3B Transparency: Coordinate-System Instability}

This section reports the Qwen2.5-3B results for transparency; they are excluded from the principal interpretation. In chat-template coordinates, the shared-direction geometry is still measurable, but it is much less one-dimensional than in Qwen2.5-7B. In raw refusal-side coordinates, the refusing baseline itself is unstable, so removal-style follow-ups can produce answer behaviour without defining a clean counterpart to the main Qwen2.5-7B/Qwen2.5-14B assay. We therefore do not use Qwen2.5-3B as principal scaling evidence.

\input{tables/qwen3b_transparency_summary}

\clearpage
\section{Scope and Robustness Analyses}

This section collects two analyses that constrain the scope of our main claims: a characterization of where non-adjacent support appears, and a follow-up that tests whether wider suppression automatically yields better refusal assembly.

\subsection{Generation-Boundary Positions}

\input{tables/refusal_token_boundary_summary}

\subsection{Wider-Patch Follow-Up}

\input{tables/refusal_wider_patch_followup}

The $-4$ slot in this follow-up lies just outside the assistant-generation boundary used in the main Qwen chat template, immediately before the \texttt{im\_start}/\texttt{assistant} boundary sequence summarized above. We did not run a full combinatorial search over skip-position patches such as $(-1,-2,-4)$. The table should therefore be read as a boundary check showing that patch width and position identity interact, not as a systematic explanation of the $-4$ reversal.

\clearpage
\section{Naturalistic Refusal Checks}\label{app:naturalistic-refusal}
\label{app:F}

\input{tables/openended_refusal}

For the naturalistic free-form follow-up, we verified the full candidate-to-retained path and treat the retained examples as a selected diagnostic set. We first established a small clean retained set under the same scoring rule used for patching and then expanded only the artifact families that continued to form stable answer-versus-refusal baselines. Email and article forms were excluded because harmful-side refusal remained unstable under the same gate, so including them would mainly add prompt-transfer noise and provide little additional support for the naturalistic claim. The retained set is intentionally diagnostic. It tests whether the directional split survives outside the multiple-choice format after strict gating. Estimating prevalence across natural harmful prompts requires a separate representative evaluation.

\begin{table}[H]
\centering
\small
\setlength{\tabcolsep}{6pt}
\begin{tabular}{lccc}
\toprule
\textbf{Artifact} & \textbf{Candidates} & \textbf{Retained} & \textbf{Retention rate} \\
\midrule
Speech & 12 & 12 & 1.00 \\
Thread & 12 & 9 & 0.75 \\
Blog & 6 & 4 & 0.67 \\
\midrule
\end{tabular}

\caption{Initial artifact-screening gate for the naturalistic free-form pipeline. These rows identify the artifact families that form stable operational-answer versus refusal baselines under the same scoring rule later used for patching.}
\label{tab:naturalistic-gate-v4}
\end{table}

\begin{table}[H]
\centering
\small
\setlength{\tabcolsep}{6pt}
\begin{tabular}{lccc}
\toprule
\textbf{Artifact} & \textbf{Candidates} & \textbf{Retained} & \textbf{Retention rate} \\
\midrule
Speech & 42 & 40 & 0.95 \\
Thread & 42 & 20 & 0.48 \\
Blog & 24 & 6 & 0.25 \\
\midrule
\end{tabular}

\caption{Naturalistic gate for the main free-form generation check in Section~\ref{sec:patching}. The retained 66-case set is usable for generation-level causal analysis, but its artifact balance remains uneven: speech prompts are much more stable than thread prompts, and blog prompts are too sparse to support a separate claim.}
\label{tab:naturalistic-gate-v5}
\end{table}

Patched generations are scored into refusal, safe-analysis, and operational-answer categories. On the main retained naturalistic set, answer release remains broad across layers 16--26 (operational answer 0.64/0.62/0.59), whereas refusal reimposition is front-loaded at layer 16 (refusal 0.45), attenuates to 0.02 by layer 22, and collapses to 0.00 by layer 26. The aggregate late-layer release-versus-refusal separation is about 0.60, with release at 0.61 and refusal at 0.008; a bootstrap interval for the late-layer gap remains strictly positive. This gap is measured on a different generation endpoint from the controlled 0.50 hidden-state locality gap, so we compare only its sign and direction, not its absolute scale. The main text therefore presents the naturalistic result as cross-setting support in a locality-restructured form, with the matched A/B experiment remaining the primary experiment.

\input{tables/naturalistic_refusal_v5}

\input{tables/naturalistic_gate_v6}

\input{tables/naturalistic_refusal_v6_stress}

We also evaluate a larger speech/thread expansion as a robustness check. It increases the retained set from 66 to 283 cases by expanding speech and thread prompts only, but that gain is highly uneven: speech retains 253 examples while thread retains only 30. Even under that stronger speech bias, the aggregate causal pattern remains stable. Answer release stays broad across layers 16--26 (operational answer 0.67/0.70/0.66), whereas refusal reimposition is front-loaded at layer 16 (refusal 0.51), attenuates to 0.02 by layer 22, and is effectively absent by layer 26 (0.01). The aggregate late-layer gap is 0.664 with a 95\% bootstrap interval of [0.613, 0.716]. The thread subset is weaker, but still directionally compatible: its late-layer gap remains positive with bootstrap interval [0.117, 0.433]. We therefore read this larger set as a speech-heavy robustness check that reinforces the same non-mirror locality profile, not as a stronger replacement for the more artifact-diverse retained set used in the main text.

%% file: tables/model_name_registry.tex
\begin{table*}[t]
\centering
\footnotesize
\setlength{\tabcolsep}{5pt}
\renewcommand{\arraystretch}{1.15}
\begin{tabular}{@{}llll@{}}
\bottomrule
\textbf{Family} & \textbf{Full model name} & \textbf{Paper shorthand} & \textbf{HF repo ID} \\
\midrule
Qwen & Qwen2.5-3B-Instruct & Qwen2.5-3B & \texttt{Qwen/Qwen2.5-3B-Instruct} \\
Qwen & Qwen2.5-7B-Instruct & Qwen2.5-7B & \texttt{Qwen/Qwen2.5-7B-Instruct} \\
Qwen & Qwen2.5-14B-Instruct & Qwen2.5-14B & \texttt{Qwen/Qwen2.5-14B-Instruct} \\
Qwen & Qwen2.5-32B-Instruct & Qwen2.5-32B & \texttt{Qwen/Qwen2.5-32B-Instruct} \\
Llama & Llama-3.1-8B-Instruct & Llama-3.1-8B & \texttt{meta-llama/Llama-3.1-8B-Instruct} \\
Llama & Llama-3.2-3B-Instruct & Llama-3.2-3B & \texttt{meta-llama/Llama-3.2-3B-Instruct} \\
Mistral & Mistral-7B-Instruct-v0.3 & Mistral-7B & \texttt{mistralai/Mistral-7B-Instruct-v0.3} \\
Gemma & Gemma-4-E4B-it & Gemma-4-E4B & \texttt{google/gemma-4-e4b-it} \\
Gemma & Gemma-4-26B-A4B-it & Gemma-4-26B-A4B & \texttt{google/gemma-4-26B-A4B-it} \\
\toprule
\end{tabular}

\caption{Model naming convention used throughout the paper. HF repo IDs are listed here for exact reference.}
\label{tab:model-name-registry}
\end{table*}

%% file: tables/backbone_extensions.tex
\begin{table*}[t]
\centering
\small

\resizebox{.7\linewidth}{!}{%
\begin{tabular}{lccc}
\toprule
\textbf{Task} & \textbf{Behaviour} & \textbf{Best decoder} & \textbf{Best bits} \\
\midrule
Withhold refusal (Llama-3.2-3B) & 83.2 refusal & mlp @ 3 & 1.000 / 1.000 \\
\midrule
\end{tabular}%
}

\caption{Appendix backbone extension retained only for the refusal mainline. Llama-3.2-3B shows the same near-tight available-but-suppressed refusal backbone at probe level. Behaviour is refusal rate. The main cross-family causal support in Appendix~\ref{app:C} uses the size-matched Llama-3.1-8B checkpoint.}
\label{tab:backbone-extensions}
\end{table*}

%% file: tables/refusal_bidirectional.tex
\begin{table*}[t]
\centering
\scriptsize

\resizebox{\linewidth}{!}{
\begin{tabular}{lcccc}
\toprule
\textbf{Direction} & \textbf{Best / plateau layers} & \textbf{Patched rate} & \textbf{Clean acc.} & \textbf{Corrupt acc.} \\
\midrule
answer $\rightarrow$ refuse & 16--26 & 1.000 top-1 recovery & 1.0 & 0.5 \\
refuse $\rightarrow$ answer & 16--27 & 0.500 plateau on the binary top-1 readout & 1.0 & 0.5 \\
\midrule
\end{tabular}}

\caption{Single-position bidirectional causal control for the main \texttt{withhold} refusal setting on public Qwen2.5-7B. The hidden-state readout is the top-1 prediction between the correct-answer token and the distractor token from the patched final-token logits. The release edit, A$\rightarrow$R (answer$\rightarrow$refuse), fully restores the hidden answer across a broad late-layer band. The reverse suppressive edit, R$\rightarrow$A (refuse$\rightarrow$answer), leaves that binary readout at the unpatched R-run baseline while still shifting the paired correct-answer-minus-distractor margin toward suppression; strong behavioral suppression appears only once earlier positions are added (Table~\ref{tab:refusal-multiposition-generation}).}
\label{tab:refusal-bidirectional}
\end{table*}

%% file: tables/refusal_generation_control.tex
\begin{table*}[t]
\centering
\scriptsize

\resizebox{\linewidth}{!}{
\begin{tabular}{lccc}
\toprule
\textbf{Direction} & \textbf{Layers tested} & \textbf{Baseline behavior} & \textbf{Best patched behavior} \\
\midrule
answer $\rightarrow$ refuse & 16, 22, 26 & refusal $1.00$, answer $0.00$ & refusal $0.50$, answer $1.00$ at layers 22/26 \\
refuse $\rightarrow$ answer & 16, 22, 26 & answer $1.00$, refusal $0.00$ & answer $0.50, 0.34, 0.00$; refusal stays $0.00$ \\
\midrule
\end{tabular}}

\caption{Single-position generation-level bidirectional patching on the main \texttt{withhold} refusal setting in public Qwen2.5-7B. The same locality gap seen in the hidden-state patch metrics survives into greedy generation: late-layer answer$\rightarrow$refuse patching releases the hidden answer locally, while the reverse refuse$\rightarrow$answer edit degrades answering without reinstating refusal. Table~\ref{tab:refusal-multiposition-generation} shows that strong behavioral answer suppression only appears once the patch widens across earlier positions.}
\label{tab:refusal-generation-control}
\end{table*}

%% file: tables/sentence_force_control.tex
\begin{table*}[t]
\centering
\small
\renewcommand{\arraystretch}{1.28}
\setlength{\tabcolsep}{4pt}

\newcommand{\metriccell}[2]{%
  \begin{minipage}[c]{\linewidth}
  \raggedright #1\\[-1pt]{\footnotesize\itshape #2}
  \end{minipage}%
}
\newcommand{\centercell}[1]{%
  \begin{tabular}[c]{@{}c@{}}#1\end{tabular}%
}
\newcommand{\headcell}[2]{%
  \begin{tabular}[c]{@{}c@{}}\textbf{#1}\\[-1pt]\textbf{#2}\end{tabular}%
}

\resizebox{\linewidth}{!}{%
\begin{tabular}{@{}
  >{\raggedright\arraybackslash}m{0.36\linewidth}
  >{\centering\arraybackslash}m{0.16\linewidth}
  >{\centering\arraybackslash}m{0.18\linewidth}
  >{\centering\arraybackslash}m{0.12\linewidth}
  >{\centering\arraybackslash}m{0.09\linewidth}
@{}}
\toprule
\multirow{3}{*}[-0.3ex]{\textbf{Metric}}
  & \multicolumn{2}{c}{\textbf{Patched behavior}}
  & \multirow{3}{*}[-0.3ex]{\textbf{Best layer(s)}}
  & \multirow{3}{*}[-0.3ex]{\textbf{Profile}} \\
\cmidrule(lr){2-3}
  & \headcell{Clean-label}{release}
  & \headcell{Override}{reimposition}
  & & \\
\midrule

\metriccell{Hidden state (A/B top-1)}
  {late-layer answer-channel switching}
  & \centercell{$\mathbf{1.000}$}
  & \centercell{$\mathbf{1.000}$}
  & \centercell{24--27}
  & \centercell{Symm.} \\

\addlinespace[5pt]

\metriccell{Generation (leading label)}
  {first-token behavioral targeting}
  & \centercell{$\mathbf{1.000}$}
  & \centercell{$\mathbf{1.000}$}
  & \centercell{26}
  & \centercell{Symm.} \\

\addlinespace[5pt]

\metriccell{Generation (exact sentence form)}
  {exact template assembly; asymmetry in opposite direction}
  & \centercell{$0.172$}
  & \centercell{$\mathbf{1.000}$}
  & \centercell{22/26 vs.\ 26}
  & \centercell{Reverse} \\

\bottomrule
\end{tabular}
}

\caption{Qwen multi-token non-refusal sentence control. The target outputs are sentence completions beginning with \texttt{A} or \texttt{B}, so the control preserves the same first-token A/B coordinate system while requiring a natural-language continuation. This makes the output format matched across directions, even though the underlying instructions are not strict semantic mirrors: the clean prompt asks for the correct label, whereas the override prompt forces the opposite label. Hidden-state switching and generation-level leading-label targeting are symmetric at late layers; only exact sentence-template reassembly is asymmetric, and that asymmetry runs in the \emph{opposite} direction from withhold refusal (favoring the override side). Profile labels: \textit{Symm.} = both directions saturate; \textit{Reverse} = asymmetric, but the suppress-side is stronger than release-side, opposite to the refusal pattern in Table~\ref{tab:refusal-core-asymmetry}.}
\label{tab:sentence-force-control}
\end{table*}

%% file: tables/out_of_set_override_controls.tex
\begin{table*}[t]
\centering
\small
\setlength{\tabcolsep}{5pt}
\begin{tabular}{lcccc}
\toprule
\textbf{Override target} & \textbf{Clean gate} & \textbf{Override gate} & \textbf{Release L22} & \textbf{Release L26} \\
\midrule
\texttt{C} & 1.00 & 1.00 & 1.00 & 1.00 \\
\texttt{PASS} & 1.00 & 1.00 & 1.00 & 1.00 \\
\texttt{NONE} & 1.00 & 1.00 & 1.00 & 1.00 \\
\texttt{SKIP} & 1.00 & 1.00 & 1.00 & 1.00 \\
\texttt{safety-(C/PASS/NONE/SKIP)} & 1.00 & 1.00 & 1.00 & 1.00 \\
\bottomrule
\end{tabular}
\caption{Out-of-set override controls in Qwen2.5-7B. The override prompt prescribes an exact target outside the A/B answer set, optionally under safety framing. Each target family uses 256 examples; the grouped safety row reports the shared value across \texttt{C}, \texttt{PASS}, \texttt{NONE}, and \texttt{SKIP}. Release columns report generation-level A$\rightarrow$R answer release under single-position patching.}
\label{tab:out-of-set-override-controls}
\end{table*}

%% file: tables/sampling_generation_controls.tex
\begin{table*}[t]
\centering
\small
\setlength{\tabcolsep}{5pt}
\begin{tabular}{lcccc}
\toprule
\textbf{Condition} & \textbf{Decoding} & \textbf{Baseline target} & \textbf{Release L22} & \textbf{Release L26} \\
\midrule
\texttt{withhold} & $T=0.7$, top-$p=0.95$ & refusal 1.00 & 1.00 & 1.00 \\
\texttt{C} & $T=0.7$, top-$p=0.95$ & override 1.00 & 1.00 & 1.00 \\
\texttt{safety-C} & $T=0.7$, top-$p=0.95$ & override 1.00 & 1.00 & 1.00 \\
\texttt{safety-PASS} & $T=0.7$, top-$p=0.95$ & override 1.00 & 1.00 & 1.00 \\
\bottomrule
\end{tabular}
\caption{Generation-level sampling robustness in Qwen2.5-7B with 10 samples per pair. Release columns report A$\rightarrow$R answer release under single-position patching. These rows measure answer-channel release under sampling.}
\label{tab:sampling-generation-controls}
\end{table*}

%% file: tables/refusal_multiposition_generation.tex
\begin{table*}[t]
\centering
\scriptsize

\resizebox{.85\linewidth}{!}{
\begin{tabular}{lccc}
\toprule
\textbf{Direction} & \textbf{Positions} & \textbf{Representative layer} & \textbf{Patched behavior} \\
\midrule
answer $\rightarrow$ refuse & $-1$ & 22/26 & refusal $0.50$, answer $1.00$ \\
answer $\rightarrow$ refuse & $(-1,-2)$ & 16 & refusal $0.00$, answer $0.50$ \\
answer $\rightarrow$ refuse & $(-1,-2,-3)$ & 16/22 & refusal $0.00$, answer $1.00$ \\
\midrule
refuse $\rightarrow$ answer & $-1$ & 16 & answer $0.50$, refusal $0.00$ \\
refuse $\rightarrow$ answer & $(-1,-2)$ & 16 & answer $0.17$, refusal $0.66$ \\
refuse $\rightarrow$ answer & $(-1,-2,-3)$ & 16 & answer $0.00$, refusal $0.84$ \\
\midrule
\end{tabular}}

\caption{Multi-position generation-level patching for the main \texttt{withhold} refusal setting in public Qwen2.5-7B. The single-position asymmetry remains the main result, but extending the patch across earlier positions strengthens both directions behaviorally. The largest gain appears in the suppressive \texttt{refuse$\rightarrow$answer} direction at layer 16: answer rate falls monotonically from 0.50 to 0.17 and then 0.00 as the patch widens from $-1$ to $(-1,-2)$ and then $(-1,-2,-3)$, while refusal rises from 0.00 to 0.66 and then 0.84. The release direction also strengthens, but remains more reliable and reaches full answer release earlier.}
\label{tab:refusal-multiposition-generation}
\end{table*}

%% file: tables/refusal_negative_controls.tex
\begin{table*}[t]
\centering
\small
\renewcommand{\arraystretch}{1.3}
\setlength{\tabcolsep}{8pt}

\resizebox{.9\linewidth}{!}{%
\begin{tabular}{@{}lccc@{}}
\toprule
 & \multicolumn{2}{c}{\textbf{Patched behavior}} & \\
\cmidrule(lr){2-3}
\textbf{Intervention} & \textbf{answer}\,$\rightarrow$\,\textbf{refuse} & \textbf{refuse}\,$\rightarrow$\,\textbf{answer} & \textbf{Effect} \\
\midrule
Single head {\footnotesize (28 heads $\times$ 3 layers)} & no release & no change (84/84) & null \\
Attention-only        & no release & no suppression & null \\
MLP-only              & no release & no suppression & null \\
\addlinespace[2pt]
\textbf{Full layer}   & \textbf{answer 1.00} & \textbf{refusal 0.84} & \textbf{full asymmetry} \\
\midrule
\multicolumn{4}{@{}l}{\textbf{Cross-family single-head suppress check (best tested head)}} \\
Qwen2.5-7B       & --- & answer $1.00$, refusal $0.00$ & null \\
Llama-3.1-8B  & --- & answer $0.83$, refusal $0.00$ & weak answer-only \\
Mistral-7B    & --- & answer $0.84$, refusal $0.00$ & weak answer-only \\
Gemma-4-E4B   & --- & answer $1.00$, refusal $0.00$ & null \\
Gemma-4-26B-A4B & --- & answer $1.00$, refusal $0.00$ & null \\
\bottomrule
\end{tabular}%
}

\caption{Negative mechanistic controls for the main \texttt{withhold} refusal site. The top block reports the original Qwen2.5-7B controls at layer 16 with positions $(-1,-2,-3)$: sparse interventions (single head, attention-only, MLP-only) produce no behavioral change in either direction, while full-layer patching reproduces the behavioral asymmetry. The lower block extends the single-head suppress check across the newly tested families by reporting the best tested head over layers 16, 22, and 26 under the same multi-position patch. Across all families, no single head reinstalls explicit refusal; at most, some heads weakly degrade answering without producing refusal strings.}
\label{tab:refusal-negative-controls}
\end{table*}

%% file: tables/cross_model_bidirectional_summary.tex
\begin{table*}[t]
\centering
\scriptsize
\setlength{\tabcolsep}{5pt}

\resizebox{\linewidth}{!}{
\begin{tabular}{lccccc}
\toprule
\textbf{Model} & \textbf{Hidden release} & \textbf{Hidden suppress} & \textbf{Gen. release} & \textbf{Gen. suppress} & \textbf{Profile} \\
\midrule
Qwen2.5-7B      & 1.00 & 0.50 & 1.00 & 0.00 & Main asym. \\
Qwen2.5-14B     & 1.00 & 0.00 & 0.00 & 0.00 & Hidden-only asym. \\
Qwen2.5-32B     & 1.00 & 0.00 & 0.00 & 0.00 & Deep hidden asym. \\
Llama-3.1-8B & 1.00 & 0.00 & 1.00 & 0.17 & Asym. \\
Mistral-7B   & 1.00 & 0.50 & 1.00 & 0.00 & Asym. \\
Gemma-4-E4B  & 1.00 & 0.00 & 1.00 & 0.00 & Maximal asym. \\
Gemma-4-26B-A4B & 1.00 & 0.00 & 1.00 & 0.00 & Maximal asym. \\
\bottomrule
\end{tabular}}

\caption{Cross-model bidirectional patching summary over the completed aligned-checkpoint runs. Hidden release and hidden suppress report the best single-position A/B readout in the answer$\rightarrow$refuse and refuse$\rightarrow$answer directions, respectively. Generation columns report the best answer release and explicit-refusal reassembly under the same single-position patching family. Qwen2.5-14B now shows a narrow late-layer hidden asymmetry: release reaches 1.00 only at layer 26, while reverse suppression stays at 0.00 and generation-level single-position patching remains inert. Qwen2.5-32B shows the same asymmetry only after shifting the hidden-state sweep to deeper layers 40--58; the earlier 16/22/26 band was a no-op. The resulting Qwen-family pattern is therefore not that the asymmetry disappears with scale, but that it survives more robustly at the hidden-state level than at the generation-control level.}
\label{tab:cross-model-bidirectional-summary}
\end{table*}

%% file: tables/cross_family_robustness_summary.tex
\begin{table*}[t]
\centering
\small
\renewcommand{\arraystretch}{1.4}
\setlength{\tabcolsep}{6pt}

\resizebox{.9\linewidth}{!}{%
\begin{tabular}{@{}lcccc@{}}
\toprule
\multirow{2}{*}[-0.35ex]{\textbf{Model}}
  & \multicolumn{2}{c}{\textbf{Removal-side behaviour}}
  & \textbf{Addition-side}
  & \multirow{2}{*}[-0.35ex]{\textbf{Profile}} \\
\cmidrule(lr){2-3}
  & \textbf{Refusal} $\downarrow$
  & \textbf{Answer} $\uparrow$
  & \textbf{Refusal assembly}
  & \\
\midrule
\multicolumn{5}{@{}l}{\textbf{Within-family scaling: asymmetry survives, control degrades}} \\
Qwen2.5-7B        & $1.000 \to 0.000$ & $0.000 \to 1.000$ & partial but effective          & Full asym. \\
\addlinespace[2pt]
\multicolumn{5}{@{}l}{\textbf{Within-family scaling: direction fragility}} \\
Qwen2.5-14B       & $1.000 \to 1.000$ & $0.000 \to 0.000$ & no-op                          & Direction inert \\
Qwen2.5-32B       & $1.000 \to 1.000$ & $0.000 \to 0.000$ & no-op                          & Direction inert \\
\addlinespace[2pt]
\multicolumn{5}{@{}l}{\textbf{Cross-family asymmetry support}} \\
Llama-3.1-8B & $1.000 \to 0.000$ & $0.000 \to 1.000$ & layer-16 saturation            & Supports \\
Mistral-7B   & $1.000 \to 0.000$ & $0.000 \to 0.500$ & layer-16 saturation            & Supports \\
Gemma-4-E4B    & $1.000 \to 0.000$ & $0.000 \to 1.000$ & fails under addition           & Maximal asym. \\
Gemma-4-26B-A4B & $1.000 \to 0.000$ & $0.000 \to 1.000$ & fails under addition           & Maximal asym. \\
\bottomrule
\end{tabular}%
}

\caption{Cross-model summary for shared-direction interventions using the paper shorthands in Table~\ref{tab:model-name-registry}. The Qwen rows show a within-family scaling gradient: the average answer-to-refusal displacement is behaviourally effective at 7B, but the same add/remove/projection family is inert at 14B and 32B even though Appendix Table~\ref{tab:cross-model-bidirectional-summary} shows that hidden-state asymmetry survives beyond 7B. The cross-family rows report checkpoints where removing the estimated displacement collapses refusal and raises answer behaviour at representative tested layers. Llama-3.1-8B is the cleanest size-matched support line, Mistral-7B is a second positive case, and the Gemma-4 rows show the most extreme removal-versus-addition split. Profile labels: \emph{Full asym.} = removal and addition together reproduce the main Qwen2.5-7B pattern; \emph{Direction inert} = concentrated geometry is present but the matched interventions are behaviourally inert; \emph{Supports} = removal collapses refusal and releases answers at representative settings; \emph{Maximal asym.} = removal is strong but addition fails to assemble refusal.}
\label{tab:cross-family-robustness-summary}
\end{table*}

%% file: tables/qwen14b_geometry_summary.tex
\begin{table*}[t]
\centering
\scriptsize
\setlength{\tabcolsep}{4pt}

\resizebox{.8\linewidth}{!}{%
\begin{tabular}{l>{\centering\arraybackslash}p{0.13\linewidth}>{\centering\arraybackslash}p{0.13\linewidth}>{\centering\arraybackslash}p{0.13\linewidth}}
\toprule
\textbf{Model} & \textbf{Layer 16} & \textbf{Layer 22} & \textbf{Layer 26} \\
\midrule
Qwen2.5-7B      & 0.849 / 4.50 & 0.841 / 4.28 & 0.891 / 4.11 \\
Qwen2.5-14B     & 0.723 / 4.62 & 0.731 / 4.58 & 0.813 / 4.75 \\
Qwen2.5-32B     & 0.875 / 3.84 & 0.906 / 4.62 & 0.875 / 4.57 \\
Llama-3.1-8B & 0.923 / 4.26 & 0.893 / 3.74 & 0.896 / 3.67 \\
Mistral-7B   & 0.928 / 4.67 & 0.925 / 4.45 & 0.917 / 4.51 \\
Gemma-4-E4B  & 0.687 / 4.84 & 0.679 / 3.22 & 0.896 / 4.32 \\
Gemma-4-26B-A4B & 0.772 / 4.60 & 0.632 / 4.40 & 0.574 / 4.51 \\
\midrule
\end{tabular}}

\caption{Cross-model geometry comparison for the shared suppression direction, using the paper shorthands in Table~\ref{tab:model-name-registry}. Each cell reports mean-direction energy share / centered effective rank at the analyzed multi-position site. Concentrated geometry is common across the tested checkpoints, but Table~\ref{tab:cross-family-robustness-summary} shows that this geometry does not by itself predict whether addition, ablation, or projection becomes a behaviorally effective refusal-control handle.}
\label{tab:qwen14b-geometry-summary}
\end{table*}

%% file: tables/qwen14b_removal_summary.tex
\begin{table*}[t]
\centering
\small
\setlength{\tabcolsep}{5pt}

\resizebox{\linewidth}{!}{%
\begin{tabular}{lccc}
\toprule
\textbf{Intervention} & \textbf{Layer 16} & \textbf{Layer 22} & \textbf{Layer 26} \\
\midrule
Addition & answer 1.00 / refusal 0.00 & answer 1.00 / refusal 0.00 & answer 1.00 / refusal 0.00 \\
Ablation & answer 0.00 / refusal 1.00 & answer 0.00 / refusal 1.00 & answer 0.00 / refusal 1.00 \\
Projection & answer 0.00 / refusal 1.00 & answer 0.00 / refusal 1.00 & answer 0.00 / refusal 1.00 \\
\midrule
\end{tabular}%
}

\caption{Shared-direction interventions in Qwen2.5-14B at $\lambda = 1.0$. Addition starts from the answer trajectory, whose baseline is answer 1.00 / refusal 0.00; ablation and projection start from the refusal trajectory, whose baseline is answer 0.00 / refusal 1.00. All three interventions are behaviorally inert at the tested late layers. Together with Table~\ref{tab:qwen14b-geometry-summary}, this makes Qwen2.5-14B a within-family geometry/control dissociation rather than a second positive transfer case.}
\label{tab:qwen14b-removal-summary}
\end{table*}

%% file: tables/llama_refusal_causal.tex
\begin{table*}[t]
\centering
\small
\renewcommand{\arraystretch}{1.3}
\setlength{\tabcolsep}{8pt}

\resizebox{.9\linewidth}{!}{
\begin{tabular}{@{}lccc@{}}
\toprule
 & \multicolumn{2}{c}{\textbf{Patched behavior}} & \\
\cmidrule(lr){2-3}
\textbf{Endpoint} & \textbf{Release} & \textbf{Suppress} & \textbf{Profile} \\
 & {\footnotesize answer\,$\rightarrow$\,refuse} & {\footnotesize refuse\,$\rightarrow$\,answer} & \\
\midrule
Two-label A/B top-1 readout              & $\mathbf{1.000}$ & $0.000$ & Asym. \\
{\footnotesize \textit{(source 1.000, target 0.664)}} & {\footnotesize recovered across tested layers} & {\footnotesize below target baseline} & \\
\addlinespace[3pt]
Generation (best layer)                  & $\mathbf{1.000}$ & $0.172$ & Asym. \\
{\footnotesize \textit{(L26 vs.\ L22)}} & {\footnotesize answer restored, refusal 0.000} & {\footnotesize refusal weakly restored; answer 0.664} & \\
\bottomrule
\end{tabular}}

\caption{Llama-3.1-8B cross-family causal support for the \texttt{withhold} refusal setting. The same bidirectional intervention remains asymmetric in a size-matched Llama checkpoint: answer release reaches the source-run ceiling, while reverse edits do not reliably reinstall explicit refusal strings. Hidden-state baselines are source $1.000$ and target $0.664$; generation baselines are answer $1.000$ / refusal $0.000$ on the A-run target and refusal $1.000$ / answer $0.000$ on the R-run target. Profile labels follow Tables~\ref{tab:refusal-core-asymmetry} and \ref{tab:nonrefusal-override-control}.}
\label{tab:llama-refusal-causal}
\end{table*}

%% file: tables/qwen3b_transparency_summary.tex
\begin{table*}[t]
\centering
\small
\setlength{\tabcolsep}{5pt}
\renewcommand{\arraystretch}{1.22}

\newcommand{\coordcell}[2]{%
  \begin{tabular}[c]{@{}l@{}}
  #1\\[-1pt]
  {\scriptsize #2}
  \end{tabular}%
}
\newcommand{\numcell}[1]{%
  \begin{tabular}[c]{@{}c@{}}
  #1
  \end{tabular}%
}

\resizebox{.84\linewidth}{!}{%
\begin{tabular}{@{}>{\raggedright\arraybackslash}
m{0.35\linewidth}
>{\centering\arraybackslash}m{0.13\linewidth}>{\centering\arraybackslash}m{0.13\linewidth}>{\centering\arraybackslash}m{0.13\linewidth}@{}}
\toprule
\textbf{Coordinate view}
  & \textbf{Layer 16}
  & \textbf{Layer 22}
  & \textbf{Layer 26} \\
\midrule
\coordcell{Chat-template geometry}{top-1 share / eff. rank}
  & \numcell{0.310 / 4.68}
  & \numcell{0.297 / 4.76}
  & \numcell{0.294 / 4.77} \\

\addlinespace[3pt]

\coordcell{Raw refusal-side removal}{best answer / refusal}
  & \numcell{1.000 / 0.000}
  & \numcell{1.000 / 0.000}
  & \numcell{1.000 / 0.000} \\
\midrule
\end{tabular}}

\caption{Qwen2.5-3B transparency summary across coordinate systems. In chat-template coordinates, the shared-direction geometry is measurable but much less concentrated than in Qwen2.5-7B or the refreshed Qwen2.5-14B run: the centered top-1 share stays near 0.30 with centered effective rank near 4.7. In the raw refusal-side coordinate system, the refusing baseline is unstable, so best tested removal runs can release answer behavior without defining a clean refusal-side counterpart to the main assay. The two views are therefore informative for transparency only, not for principal scaling claims.}
\label{tab:qwen3b-transparency-summary}
\end{table*}

%% file: tables/refusal_token_boundary_summary.tex
\noindent\begin{minipage}{\linewidth}
\centering
\small
\setlength{\tabcolsep}{6pt}

\resizebox{\linewidth}{!}{%
\begin{tabular}{lccc}
\toprule
\textbf{Model} & \textbf{Position $\textbf{-1}$} & \textbf{Position $\textbf{-2}$} & \textbf{Position $\textbf{-3}$} \\
\midrule
Qwen2.5-7B & newline & \texttt{assistant} & \texttt{<|im\_start|>} \\
Llama-3.2-3B & double newline & \texttt{<|end\_header\_id|>} & \texttt{assistant} \\
\midrule
\end{tabular}%
}

\captionsetup{hypcap=false}
\captionof{table}{Generation-boundary positions carrying the strongest non-adjacent support in the tested models. In both Qwen2.5-7B and Llama-3.2-3B, the late-layer position-$-2$ and position-$-3$ effects fall on structural boundary tokens rather than on simple lexical cue words. This supports the narrower claim used in the main text: non-adjacent support concentrates at generation-boundary positions and is not reducible to simple keyword anchoring.}
\label{tab:refusal-token-boundary-summary}
\end{minipage}

%% file: tables/refusal_wider_patch_followup.tex
\noindent\begin{minipage}{\linewidth}
\centering
\small
\setlength{\tabcolsep}{5pt}

\resizebox{\linewidth}{!}{%
\begin{tabular}{lcc}
\toprule
\textbf{Patch extent} & \textbf{Layer 16 (answer / refusal)} & \textbf{Layer 22 (answer / refusal)} \\
\midrule
$-1$ & 0.50 / 0.00 & 0.34 / 0.00 \\
$(-1,-2)$ & 0.17 / 0.66 & 0.00 / 0.00 \\
$(-1,-2,-3)$ & 0.00 / 0.84 & 0.00 / 0.17 \\
$(-1,-2,-3,-4)$ & 1.00 / 0.00 & 0.00 / 0.00 \\
\midrule
\end{tabular}%
}

\captionsetup{hypcap=false}
\captionof{table}{Wider-patch follow-up in Qwen2.5-7B for the suppressive \texttt{refuse$\rightarrow$answer} direction. Extending the patch from $(-1,-2,-3)$ to $(-1,-2,-3,-4)$ does not improve refusal assembly: at layer 16 it fully reverses the earlier gain, restoring answer to 1.00 and refusal to 0.00, while at layer 22 it still suppresses answers without reinstating strong refusal. We treat this as a position-combination boundary or interference case, not as evidence for a monotone wider-is-better rule. The result supports the main-text reading that stronger suppression does not automatically yield more reliable refusal assembly.}
\label{tab:refusal-wider-patch-followup}
\end{minipage}

%% file: tables/openended_refusal.tex
\begin{table*}[t]
\centering
\small

\resizebox{\linewidth}{!}{%
\begin{tabular}{lccccc}
\toprule
\textbf{Model} & \textbf{Refusal rate (\%)} & \textbf{Best linear (\%)} & \textbf{Best MLP (\%)} & \textbf{Best tuned lens (\%)} & \textbf{3-seed linear (\%)} \\
\midrule
Qwen2.5-3B & 87.0 & 52.0 & 60.0 & 52.0 & $57.3\pm4.6$ \\
Qwen2.5-7B & 99.0 & 52.0 & 60.0 & 56.0 & $53.3\pm2.3$ \\
\bottomrule
\end{tabular}}

\caption{Open-ended refusal extension on public harmful prompts from JailbreakBench (JBB-Behaviors; \citealp{chao2024jailbreakbench}). Labels are balanced ten-way harm categories. This provides a weaker but more realistic refusal target relative to the synthetic multiple-choice benchmark. Even so, the completed public-Qwen checkpoints still refuse strongly while retaining substantial recoverable harm-category signal in the final-token hidden state.}
\label{tab:openended-refusal}
\end{table*}

%% file: tables/naturalistic_refusal_v5.tex
\begin{table}[t]
\centering
\small
\setlength{\tabcolsep}{6pt}

\begin{tabular}{lcc}
\toprule
\textbf{Layer} & \textbf{A$\rightarrow$R: operational answer} & \textbf{R$\rightarrow$A: refusal} \\
\midrule
16 & 0.64 & 0.45 \\
22 & 0.62 & 0.02 \\
26 & 0.59 & 0.00 \\
\midrule
\end{tabular}

\caption{Generation-level naturalistic bidirectional patching on the retained Qwen2.5-7B-Instruct free-form set (108 candidate pairs, 66 retained across speech, thread, and blog). This retained set serves as a diagnostic cross-setting evaluation; prevalence estimation would require representative sampling. The directional asymmetry persists in weaker, locality-restructured form: answer release remains broad across late layers, whereas behavioral refusal reimposition is front-loaded at layer 16 and near-zero by layer 26.}
\label{tab:naturalistic-refusal-v5}
\end{table}

%% file: tables/naturalistic_gate_v6.tex
\begin{table}[t]
\centering
\small
\setlength{\tabcolsep}{6pt}
\begin{tabular}{lccc}
\toprule
\textbf{Artifact} & \textbf{Candidates} & \textbf{Retained} & \textbf{Retention rate} \\
\midrule
Speech & 336 & 253 & 0.75 \\
Thread & 98 & 30 & 0.31 \\
\midrule
\end{tabular}

\caption{Expanded speech/thread naturalistic gate used for the appendix robustness check. The retained set grows to 283 cases, but the expansion is driven primarily by speech prompts while thread retention remains much weaker.}
\label{tab:naturalistic-gate-v6}
\end{table}

%% file: tables/naturalistic_refusal_v6_stress.tex
\begin{table}[t]
\centering
\small
\setlength{\tabcolsep}{6pt}
\begin{tabular}{lcc}
\toprule
\textbf{Layer} & \textbf{A$\rightarrow$R: operational answer} & \textbf{R$\rightarrow$A: refusal} \\
\midrule
16 & 0.67 & 0.51 \\
22 & 0.70 & 0.02 \\
26 & 0.66 & 0.01 \\
\midrule
\end{tabular}

\caption{Speech-heavy naturalistic robustness check (434 candidate pairs, 283 retained). The aggregate result is directionally consistent with the main 66-case set: answer release remains broad across late layers, whereas behavioral refusal reimposition is front-loaded and near-zero by layers 22--26.}
\label{tab:naturalistic-refusal-v6-stress}
\end{table}